\pdfoutput=1
\documentclass{article}
\usepackage{ijcai25}

\usepackage{times}
\usepackage{soul}
\usepackage{url}
\usepackage[hidelinks]{hyperref}
\usepackage[utf8]{inputenc}
\usepackage[small]{caption}
\usepackage{graphicx}
\usepackage{amsmath}
\usepackage{amsthm}
\usepackage{amssymb}
\usepackage{tabularx}
\usepackage{booktabs}
\usepackage{multirow}
\usepackage[table,xcdraw]{xcolor}
\usepackage{algorithm}
\usepackage{algpseudocode}
\usepackage[switch]{lineno}

\title{Bi-directional Curriculum Learning for Graph Anomaly Detection: Dual Focus on Homogeneity and Heterogeneity}

\author{
Yitong Hao$^1$
\and
Enbo He$^1$\and
Yue Zhang$^1$\And
Guisheng Yin$^1$\\
\affiliations
$^1$College of Computer Science and Technology, Harbin Engineering Universitym, Harbin, China\\
\emails
\{haoyitong, enochhe\}@hrbeu.edu.cn,
zycg87@sina.com,
yinguishengabc@163.com
}
\begin{document}

\maketitle

\begin{abstract}
Graph anomaly detection (GAD) aims to identify nodes from a graph that are significantly different from normal patterns.  Most previous studies are model-driven, focusing on enhancing the detection effect by improving the model structure. However, these approaches often treat all nodes equally, neglecting the different contributions of various nodes to the training. Therefore, we introduce graph curriculum learning as a simple and effective plug-and-play module to optimize GAD methods. The existing graph curriculum learning mainly focuses on the homogeneity of graphs and treats nodes with high homogeneity as easy nodes. In fact, GAD models can handle not only graph homogeneity but also heterogeneity, which leads to the unsuitability of these existing methods. To address this problem, we propose an innovative \textbf{B}i-directional \textbf{C}urriculum \textbf{L}earning strategy (BCL), which considers nodes with higher and lower similarity to neighbor nodes as simple nodes in the direction of focusing on homogeneity and focusing on heterogeneity, respectively, and prioritizes their training. Extensive experiments show that BCL can be quickly integrated into existing detection processes and significantly improves the performance of ten GAD anomaly detection models on seven commonly used datasets.
\end{abstract}

\section{Introduction}
The objective of Graph Anomaly Detection (GAD) is to identify anomaly nodes that are significantly different from the regular patterns within the graph\cite{asurvey}. The identification of these anomaly nodes is of great significance for maintaining social welfare and security and has a wide range of applications including the identification of fake news in social media\cite{fakenews}, the discovery of rare molecular drugs\cite{drug}, and brain health monitoring\cite{health}. These anomalous nodes are not only scarce but also adept at concealing themselves among normal nodes, thereby increasing the complexity and difficulty of detection.

With the advancement of deep learning, numerous studies in the field of GAD have emerged, most of which are model-driven and focus on improving model architectures to enhance detection performance\cite{deepgadsurvey}. However, these methods typically treat all nodes equally, ignoring the varying contributions of different nodes to the training process. Unlike these model-driven approaches, our research focuses on leveraging the data itself to improve performance. Therefore, we introduce Graph Curriculum Learning (GCL) to optimize existing anomaly detection methods. 

\begin{figure}[htbp]
	\centering
	\includegraphics[width=1\linewidth]{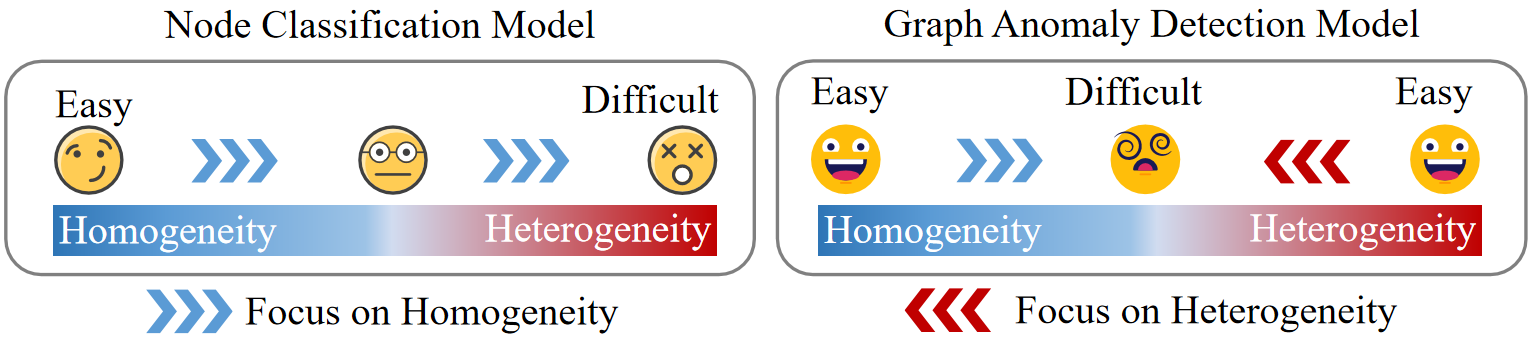} 
	\vspace{-10pt}
	\caption{The Impact of Training Scheduler}
	\label{fig:1}
\end{figure}

Graph Curriculum Learning can accelerate the training of machine learning models and enhance their generalization and detection accuracy\cite{CLNode,RCL}. However, existing GCL methods fail to fully leverage their potential when applied to GAD models, as illustrated in Figure \ref{fig:1}. Specifically, as shown in the left subplot of Figure \ref{fig:1}, GCL methods are primarily designed for node classification models that handle the homogeneity of graphs. Thus, prioritizing nodes with high homogeneity as easy samples for training effectively improves the performance of node classification models. However, as shown in the right subplot of Figure \ref{fig:1}, GAD models not only handle homogeneity but also heterogeneity. Therefore, for GAD models, nodes with both high homogeneity and high heterogeneity should be considered easy samples. Existing GCL methods overlook this aspect, resulting in suboptimal performance when applied to GAD models.

To address this challenge, we propose a novel \textbf{B}i-directional \textbf{C}urriculum \textbf{L}earning (BCL) strategy, designed to optimize training strategies tailored to the characteristics of GAD models and enhance their performance. We introduce a bi-directional difficulty score calculation method. Nodes are sorted based on this score, with low-score nodes considered easy samples in the homogeneity direction and high-score nodes considered easy samples in the heterogeneity direction. These easy samples are prioritized for training, with difficult samples gradually introduced into the GAD model according to a pacing function. Finally, the information from both directions is integrated to form the final anomaly detection results.
The main contributions of this study include:
\begin{itemize}
	\item \textbf{A Curriculum Learning Strategy for GAD:} A curriculum learning approach is introduced for the first time as a plug-and-play module in Graph Anomaly Detection. This strategy optimizes the training process based on the characteristics of GAD models, thereby enhancing existing anomaly detection methods.
	\item \textbf{A Difficulty Score Calculation Method:} A simple and efficient method for calculating bi-directional difficulty scores is proposed. This method quantifies the difficulty of nodes in terms of both homogeneity and heterogeneity, providing two directional training pathways for GAD models.
	\item \textbf{Empirical Results:} Extensive experiments on seven graph datasets validate that the proposed method significantly improves the performance of ten anomaly detection models, and further analysis of the six research questions provides a deeper perspective on our proposed method.
\end{itemize}

\section{Related Work}
\subsection{Supervised Graph Anomaly Detection}
Deep learning, particularly Graph Neural Networks (GNNs), has made significant strides in GAD by capturing both attribute and structural information to identify anomalies. Various GNN-based models have been developed for the challenges faced in this field, including efficiently capturing graph structure information and solving the problem of imbalanced label distribution. Recent advancements include methods like U-A2GAD\cite{UA2GAD}, which combines spectral methods, kNN, and kFN, and employs attention mechanisms to enhance detection performance. CARE-GNN\cite{caregnn} enhances fraud detection by improving the aggregation process of GNNs against disguised fraudsters. SplitGNN\cite{splitgnn} addresses fraud detection against heterogeneity by splitting a graph's heterogeneous and homogeneous edges. PC-GNN\cite{pcgnn} tackles class imbalance by using a label balance sampler and a learnable parameterized distance function. These models represent the evolution of graph anomaly detection, focusing on different aspects from node and edge anomalies to subgraph and entire graph anomalies.
\subsection{Curriculum Learning}
Graph Curriculum Learning applies the educational concept to the realm of graph neural networks, enhancing their ability to generalize and robustly handle noisy data. CLNode\cite{CLNode} introduces a selective training strategy for GNNs, prioritizing nodes based on a multi-perspective difficulty measurer to mitigate the impact of low-quality training nodes. CuCo\cite{cuco} presents a self-supervised framework that employs curriculum contrastive learning to sort negative samples by difficulty, optimizing graph representations through a scoring function and pacing function. CurGraph\cite{CurGraph} utilizes infomax to obtain graph embeddings and models their distributions with a neural density estimator. It calculates difficulty scores based on intra-class and inter-class distributions, facilitating a smooth transition from easy to hard samples. RCL\cite{RCL} addresses the challenge of learning dependencies in graph-structured data by gradually incorporating edges based on their difficulty, quantified through a self-supervised approach, and ensuring numerical stability through an edge reweighting scheme. 
\section{PRELIMINARIES}
\textbf{Definition 1.} \textbf{Attributed Graph:} Given a attributed graph $\mathcal{G} =\left ( \mathcal{V}, \mathcal{E}, \mathbf{X} \right) $, where $\mathcal{V} =\left\{ v_1, \cdots, v_n \right\}$ is the set of nodes, the number of nodes $|\mathcal{V} |$ is $N$, and the $\mathcal{E} =\left\{ e_1, \cdots , e_m \right\}  $ is the set of edges and the number of edges $|\mathcal{E} |$ is $m$.  $\mathbf{A}\in \mathbb{R} ^{n\times n}$ is the adjacency matrix without self-loop, where $n$ is the number of nodes. If there is a edge connecting the $i$-th and $j$-th node, $\mathbf{A}_{i,j}=1$. Otherwise, $\mathbf{A}_{i,j}=0$. $\mathbf{X}\in \mathbb{R} ^ {n\times f}$ is the node attribute matrix, where $f$ is the dimension of the attribute vector.
\textbf{Definition 2.} \textbf{Graph Anomaly Detection:} For a attributed graph $\mathcal{G} =\left ( \mathcal{V}, \mathcal{E}, \mathbf{X} \right) $, our aim is to learn a scoring function $score\left( \cdot \right) $ for qualifying the degree of abnormality. To be specific, the larger the anomaly score indicates the node is more likely to be an anomaly.
\textbf{Definition 3.} \textbf{Curriculum Learning:} Curriculum learning reduces the impact of low-quality samples by training models with a sequence of gradually more difficult subsets $<Q_1, ...,Q_t,...,Q_T>$ over $T$ training epochs. Each criterion $Q_t$ starts with easy samples and progressively includes harder ones. This approach requires designing a difficulty measure and a training scheduler to generate these subsets based on node difficulty.
\section{METHODOLOGY}
\begin{figure*}[htbp]
	\centering
	\includegraphics[width=0.9\linewidth]{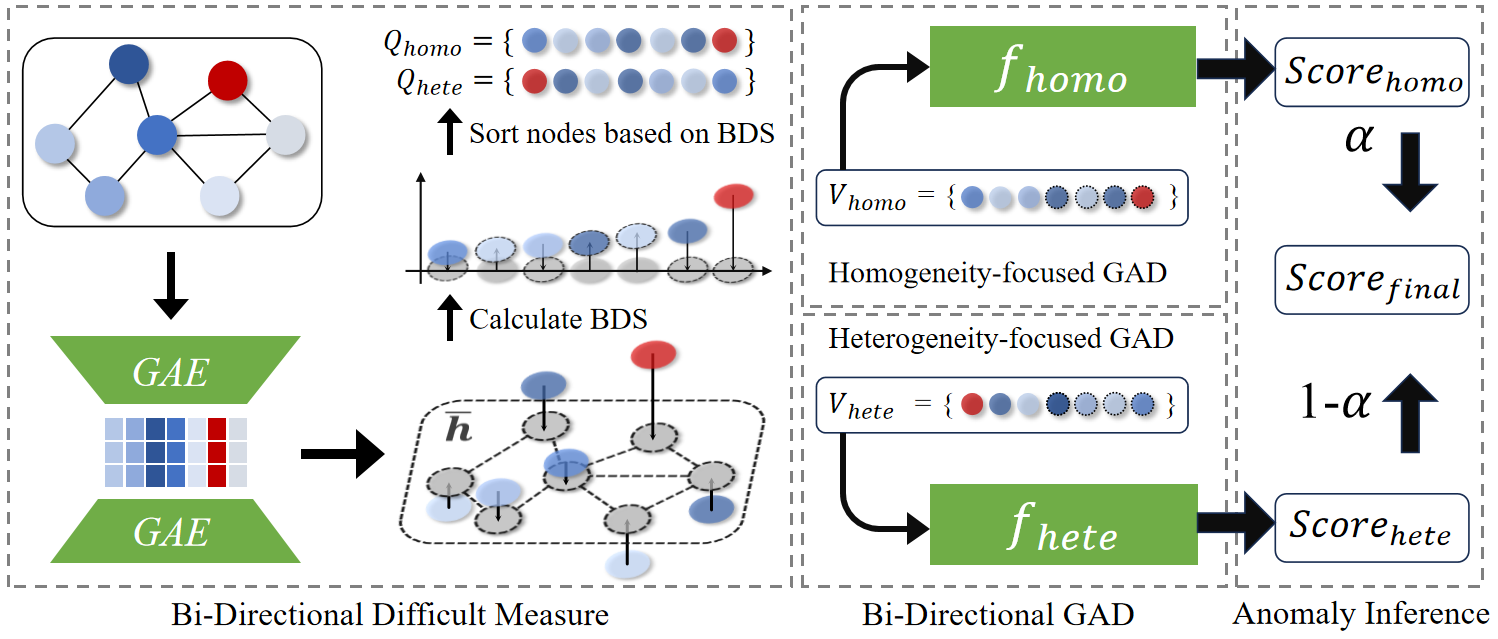} 
	\caption{The overall framework of BCL}
	\label{fig:2}
\end{figure*}
The design details of BCL will be introduced in this section. The overall framework of BCL is illustrated in Figure \ref{fig:2}, and the process of BCL is detailed in Algorithm \ref{algorithm 1}.

\subsection{Bi-Directional Difficult Measure}
A novel method for quantifying node difficulty in GAD tasks is introduced, termed the Bi-directional Difficulty Measure. This method serves as the first key component of the BCL framework. Existing graph curriculum learning methods typically rely on the label distribution of neighboring nodes, which performs well in multi-class node classification tasks. However, in GAD tasks, nodes are categorized into only two classes: normal and anomalous. Anomalous nodes often have predominantly normal neighbors, leading to similar label distributions as normal nodes, making it difficult to distinguish between them. Moreover, anomalous nodes themselves exhibit little similarity, precluding the use of homogeneity-based methods for identification.

To address these challenges, a method to quantify the homogeneity and heterogeneity of nodes is proposed, and training strategies are designed accordingly. The specific steps are as follows:

The training graph fed into the Graph Autoencoder, and the learning objective can be presented as follows:
\begin{eqnarray}
	\label{}
	\mathbf{w}_{pre}^{(t)} = \arg\min_{W} \mathcal{L}(f(\mathbf{X};\mathbf{W}), \mathbf{X})
\end{eqnarray}
A two-layer GCN is used as an encoder to aggregate neighborhood information for each node and obtain the $i$-th node representations $h_i$:
\begin{eqnarray}
	\label{}
	h_i=GCN(\mathbf{X}_i, \mathbf{A};\mathbf{W})
\end{eqnarray}
Next, the global mean of all node representations $\overline{h}$  is computed as a baseline, which reflects the overall feature distribution of nodes in the graph. The difference between each node's representation and the global mean is then calculated to obtain the bi-directional difficulty score ($BDS$):
\begin{eqnarray}
	\label{}
	BDS(v_i)=\| h_i - \overline{h} \|_1= \|  h_i - \frac{1}{N} \sum_i^N h_i \|_1
\end{eqnarray}
where $||\cdot||_1$ denotes the Euclidean norm, and  $N$ is the total number of nodes in the graph. Nodes with high homogeneity, which share similar features and behaviors with the majority of nodes, have representations $h_i$ close to $\overline{h}$, resulting in lower $BDS$. Conversely, nodes with high heterogeneity, which significantly differ from the majority, have larger differences, leading to higher $BDS$. This approach provides an intuitive measure of each node's deviation from the overall feature distribution, facilitating efficient and cost-effective quantification of node difficulty.

The nodes are sorted in ascending order of $BDS$, such that $v_{i1},v_{i2},\dots,v_{{iN}}$ satisfy:
\begin{eqnarray}
	\label{}
	BDS(v_{i1}) \leqslant BDS(v_{i2}) \leqslant \dots \leqslant BDS(v_{iN})
\end{eqnarray}

This sequence forms the homogeneity sequence $Q_{homo}=\{ v_{i1}, v_{i2},\dots, v_{iN} \} $, enabling the model to prioritize learning nodes with high homogeneity in the early stages of training. Leveraging the model's ability to handle both homogeneity and heterogeneity in graphs, nodes are also sorted in descending order of their difficulty scores to form the heterogeneity sequence $ Q_{hete} = \{ v_{iN}, \dots, v_{i2}, v_{i1} \} $. This sequence prioritizes nodes with high heterogeneity, which are typically easier for anomaly detection models to identify.

\begin{figure}[htbp]
	\centering
%	\vspace{-10pt}
	\includegraphics[width=1\linewidth]{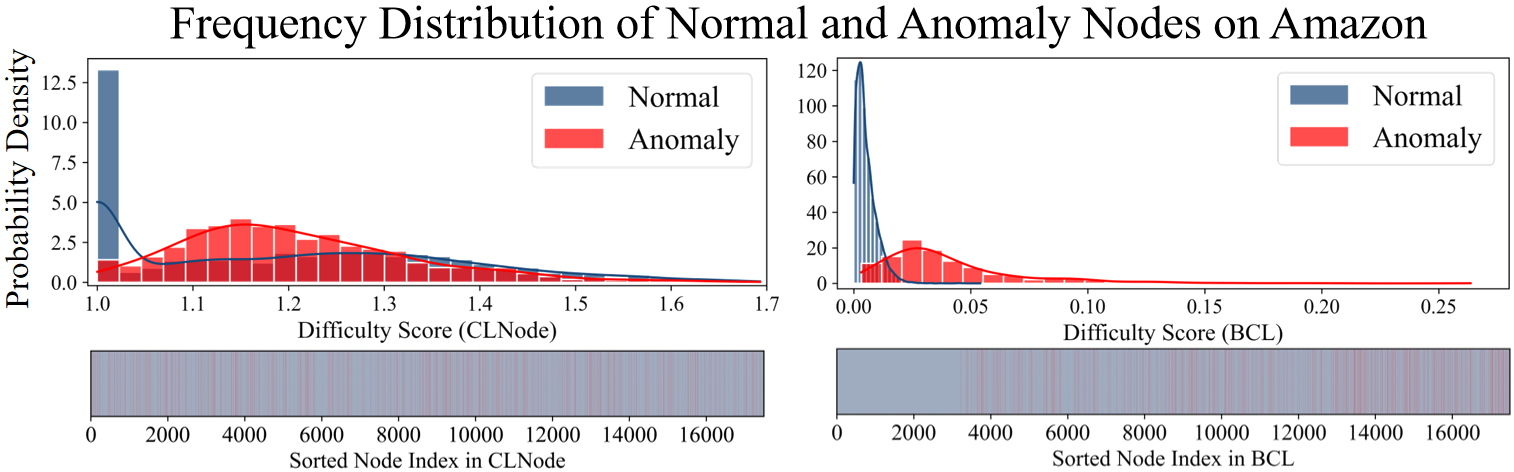} 
	\vspace{-10pt}
	\caption{Comparison of Difficult Measure for CLNode and BCL}
	\label{fig:3}
%	\vspace{-10pt}
\end{figure}

Figure \ref{fig:3} presents the difficult measure methods of CLNode and BCL on the Amazon. The upper two subplots illustrate the probability density distributions of normal and anomalous nodes under the two methods, while the lower two subplots show the sorted results after calculating the difficulty scores, with red lines indicating anomalous nodes.
It can be observed that CLNode has limitations in distinguishing between normal and anomalous nodes. The difficulty score distribution of CLNode shows significant overlap between normal and anomalous nodes, making it difficult to effectively differentiate between the two. Additionally, there is no clear separation of normal and anomalous nodes in the sorted sequence.

In contrast, BCL is more effective in identifying anomalous nodes and distinguishing them from normal nodes. The probability distributions of normal and anomalous nodes are distinctly different under BCL, with normal nodes concentrating in the lower difficulty score range and anomalous nodes in the higher range. After sorting the nodes by $BDS$, anomalous nodes are found to have relatively higher scores and are predominantly located in the latter half of the sequence. This indicates that BCL's difficulty score calculation and sorting can better capture the characteristics of anomalous nodes, thereby improving anomaly detection performance.
\begin{algorithm}[t]
	\caption{BCL} 
	\begin{algorithmic}[1]
		\Require A graph $\mathcal{G} = (\mathcal{V}, \mathcal{E}, \mathbf{X})$, the train node set $\mathcal{V}_{train}$, the input labels $Y$, the GAE model $f_{pre}$, the GAD model $f_{homo}$ focus on homogeneity, the GAD methods focus on heterogeneity $f_{hete}$, the hyper-parameters $\alpha$, $\lambda_0^{homo}$, $T_{homo}$, $\lambda_0^{hete}$, $T_{hete}$;
		\Ensure
		The anomaly score: $Score_{final}$;
		\State Initialize $\mathbf{W}_{pre}^{(0)}$;
		\While{Not converge}
		\State $\mathbf{W}_{pre}^{(t)} = \arg\min_{W} \mathcal{L}(f(\mathbf{X};\mathbf{W}), \mathbf{X})$;
		\EndWhile
		\For {$v_i \in \mathcal{V}$}
		\State Calculate bi-directional difficulty score $\text{BDS}(v_i)$;
		\EndFor
		\State Sort nodes in ascending order of $BDS(v_i)$ gets $Q_{homo}$;
		\State Initialize parameters of GAD model $f_{homo}$;
		\State Let $t = 1$;
		\While{ $t < T_{homo}$ or not converge}
		\State $\lambda^{homo}_{t} = g(t)$;
		\State $ \mathcal{V}_{homo} \leftarrow \{ \mathcal{V}_{train}[i] \mid i \in Q_{homo}[1:\lambda^{homo}_{t} \times |\mathcal{V}|] \}$;
		\State Use $f_{homo}$ predict the label $Y_{homo}$;
%		\State $\mathcal{L}_{homo}( f_{homo}(\mathbf{X},\mathbf{A};\mathbf{w}), Y)$;
		\State Calculate $\mathcal{L}_{homo}$ on $ \{ Y_{homo}[v], Y[v] \mid v \in  \mathcal{V}_{train}\}$;
		\State Back-propagation on $f_{homo}$ for minimizing $\mathcal{L}_{homo}$;
		\State $t \leftarrow t+1$;
		\EndWhile
		\State Sort nodes in descending order of $BDS(v_i)$ gets $Q_{hete}$;
		\State Initialize parameters of GAD model $f_{hete}$;
%		\State Let $t = 1$;
%		\While{ $t < T_{hete}$ or not converge}
%		\State $\lambda^{hete}_{t} = g(t)$;
%		\State $ \mathcal{V}_{hete} \leftarrow \{ \mathcal{V}_{train}[i] \mid i \in Q_{hete}[1:\lambda^{hete}_{t} \times |\mathcal{V}|] \}$;
%		\State Use $f_{hete}$ predict the label $Y_{hete}$;
%		\State Calculate $\mathcal{L}_{hete}$ on $ \{ Y_{hete}[v], Y[v] \mid v \in  \mathcal{V}_{train}\}$;
%		\State Back-propagation on $f_{hete}$ for minimizing $\mathcal{L}_{hete}$;
%		\State $t \leftarrow t+1$;
%		\EndWhile
		\State Similarly, train $f_{hete}$ using the same procedure as $f_{homo}$ but with $\lambda^{hete}_{t}$ and $T_{hete}$;
		\State Calculate the final anomaly score: 
		\State $Score_{final}= \alpha * Score_{homo} + (1 - \alpha) * Score_{hete}$	;	
	\end{algorithmic}
	\label{algorithm 1}
\end{algorithm}

\subsection{Bi-directional Continuous Training Scheduler}
To fully leverage the capability of GAD models to handle both homogeneity and heterogeneity simultaneously, two identical GAD models are employed, each focusing on one of these aspects. This dual-focus strategy ensures that both models start training from easy samples and gradually transition to more complex ones.

%After measuring the Bi-directional difficulty score of the nodes, we designed a Continuous Training Scheduler to generate easy to difficult curriculum. Dynamically adjust the order of training samples based on the $BDS$ to train a better GAD model.
%To be more specific, we first sort the training set $\mathcal{V}_{train}$ according to the order of $Q_{homo} $ or $Q_{hete} $ to obtain $\mathcal{V}_{homo} $ or $\mathcal{V}_{hete}$, ensuring that the model can learn from the easiest samples.
\subsubsection{Homogeneity-focused GAD Model}
The GAD model focusing on homogeneity, denoted as $f_{homo}$, prioritizes training nodes with lower $BDS$. It initially concentrates on nodes with high homogeneity—those whose features and behaviors are similar to the majority of nodes. By prioritizing these nodes, the model rapidly establishes a foundational understanding of the normal patterns within the graph data.

During training, the training set $\mathcal{V}_{train}$ is filtered according to the order of $Q_{homo}$ to obtain $\mathcal{V}_{homo}$, which can be expressed as:
\begin{eqnarray}
	\label{}
	\mathcal{V}_{homo} = \{\mathcal{V}_{train}[i] \mid i \in Q_{homo}[1:\lambda_{t}^{homo} \cdot |\mathcal{V}|]\}
\end{eqnarray}
where $\lambda_t^{homo}$ is a parameter that gradually increases training epochs, controlling the size of the training subset $\mathcal{V}_{homo}$.
\subsubsection{Heterogeneity-focused GAD Model}
Conversely, the GAD model focusing on heterogeneity, denoted as $f_{hete}$, prioritizes nodes with higher BDS values. These nodes, which significantly differ from the majority, are potential anomalies. By prioritizing these nodes, the model quickly identifies anomalous patterns.
During training, the training set $\mathcal{V}_{train}$ is filtered according to the order of $Q_{hete}$ to obtain $\mathcal{V}_{hete}$, which can be expressed as:
\begin{eqnarray}
	\label{}
	\mathcal{V}_{hete} = \{\mathcal{V}_{train}[i] \mid i \in Q_{hete}[1:\lambda^{hete}_{t} \cdot |\mathcal{V}|]\}
\end{eqnarray}
where $\lambda^{hete}_{t}$ is a parameter similar to $\lambda^{homo}_{t}$, controlling the size of the training subset $\mathcal{V}_{hete}$.

\subsubsection{Continuous Training Scheduler}
To achieve a smooth transition from easy to difficult samples, a continuous training scheduler is introduced. Specifically, a pacing function $g(t)$, maps each training epoch $t$ to a scalar $\lambda_{t} in (0,1\rceil$, representing the proportion of the easiest nodes available for training at epoch $t$. $\lambda_0$ denotes the initial proportion of easy nodes, and $T$ is the epoch when $g(t)$ first reaches 1. Three pacing functions are considered to control the rate at which difficult samples are introduced:
\begin{eqnarray}
	\label{}
	linear: g(t)=min(1, \lambda_{0} + ( 1 - \lambda_{0} * \frac{t}{T})
\end{eqnarray}
\begin{eqnarray}
	\label{}
	root: g(t) = min(1,\sqrt{(\lambda_{0})^2 ~~+  [1 - (\lambda_{0})^2] * \frac{t}{T}} )
\end{eqnarray}
\begin{eqnarray}
	\label{}
	geometric: g(t) = min(1, 2^{log_{2}\lambda_{0}-log_2\lambda_{0} * \frac{t}{T}})
\end{eqnarray}
These pacing functions introduce difficult samples at different rates. The linear function uniformly increases the difficulty of training nodes over time, the root function introduces more difficult nodes more quickly, and the geometric function focuses more on easy nodes initially.
The BCL framework leverages these pacing functions to continuously introduce training nodes into the process, assigning appropriate training weights based on node difficulty. Specifically, more difficult nodes are introduced later in the training process, meaning they have smaller training weights. This strategy enables the model to quickly learn easily recognizable samples in the early stages and gradually focus on more challenging samples later on.

When $t=T$, training does not stop immediately. Instead, the entire training set is used to continue training the GAD model until convergence on the validation set. This continuous training strategy ensures that the model fully utilizes all available training data while maintaining sensitivity to anomalous nodes.

By employing this bi-directional continuous training scheduler, the BCL framework optimizes the training process from both homogeneity and heterogeneity perspectives and further enhances the model's ability to detect anomalous nodes by integrating results from both directions.
\subsection{Bi-directional Fusion and Anomaly Inference}
Through the BCL strategy, the GAD model is trained from both homogeneity and heterogeneity perspectives. While the GAD models in these two directions are trained independently, their output results are combined to form the final anomaly detection result. The formula for this combination is as follows:
\begin{eqnarray}
	\label{}
	Score_{final}=\alpha Score_{homo}+(1-\alpha)Score_{hete}
\end{eqnarray}
where $\alpha$ is a weighting parameter to balance the contributions of the homogeneity and heterogeneity scores. This fusion approach not only integrates information from both directions but also adapts to the characteristics of different graph datasets and anomaly patterns. The composite anomaly score for each node reflects its deviation from the normal pattern. By considering the performance of nodes on both homogeneity and heterogeneity dimensions, this composite-score-based method enhances the accuracy of anomaly detection.
\subsection{Complexity Analysis}
The overall time complexity of the proposed Bi-directional Curriculum Learning (BCL) framework is $O(E N d) + O(NlogN)$, where $E$ is the number of training epochs, $N$ is the number of nodes in the graph, and $d$ is the dimension of the node features. This complexity primarily stems from the training of the Graph Autoencoder, which involves forward and backward passes with a time complexity of $O(N d)$. Additionally, calculating node representations and sorting nodes based on their difficulty scores contribute $O(N d)$ and $O(N log N)$, respectively. The iterative training of the GAD models in both homogeneity and heterogeneity directions adds $ O(E N d)$, with the fusion step being $O(N)$. The dominant term, $O(E N d)$, reflects the iterative nature of the training process, while sorting contributes the additional $O(N log N)$ complexity. This indicates that BCL is efficient and well-suited for graph anomaly detection tasks.

\section{Experiment}
In this section, we conduct extensive experiments to seek answers to the following research questions:

\textbf{RQ1:} How effective is the introduction of Curriculum Learning in improving the performance of existing anomaly detectors? 

\textbf{RQ2:} How well does the BCL framework compare to current state-of-the-art Curriculum Learning training strategies for graph anomaly detection tasks?

\textbf{RQ3:} Why is it important to focus on both homogeneity and heterogeneity rather than just one direction?

\textbf{RQ4:} How does the Training Scheduler influence the performance gains?

\textbf{RQ5:} How does balancing parameter $\alpha$, which balances the two-direction model, affect the BCL framework?

\textbf{RQ6:} How the hyper-parameters $\lambda_0$ and $T$ affect the performance of BCL?

\subsection{Experimental Setups}
\subsubsection{Dataset.} 
We incorporate seven extensively utilized graph anomaly detection datasets for comprehensive evaluation. These datasets encompass Amazon, Yelp\cite{yelpchi}, Reddit\cite{reddit}, Weibo, Facebook, Elliptic\cite{elliptic}, and Questions. The detailed statistical summary of these seven datasets is presented in Table \ref{Table 1}.

\begin{table}[h]
	\centering
	
%	\vspace{-10pt}
	\setlength{\tabcolsep}{3pt}
	\scriptsize
	\begin{tabular}{c|cccccc}
		\hline
		& \#Nodes & \#Edges & \#Feat.   & \#Anomaly & \#Train & \#Feature Type   \\ \hline
		Amazon    & 11,944  & 4,398,392 & 25      & 9.50\%  & 40\%  & Misc. Information \\
		YelpChi   & 45,954  & 3,846,979 & 32      & 14.50\% & 40\%  & Misc. Information \\ 
		Reddit    & 10,984  & 168,016   & 64      & 3.30\%  & 40\%  & Text Embedding    \\
		Weibo     & 8,405   & 407,963   & 400     & 10.30\% & 40\%  & Text Embedding    \\
		Facebook  & 1081    & 27,552    & 576     & 2.30\%  & 40\%  & Text Embedding    \\
		Elliptic  & 203,769 & 234,355   & 166     & 9.80\%  & 40\%  & Misc. Information \\
		Questions & 48,921  & 153,540   & 301     & 3.00\%  & 40\%  & Text Embedding    \\ \hline
	\end{tabular}
	\caption{The detailed statistical summary of these seven datasets}
	\label{Table 1}
\end{table}

\begin{table*}[t]
	\centering
	\setlength{\tabcolsep}{7pt}
	\setlength{\extrarowheight}{0pt}
	
%	\scriptsize
	\tiny
	\vspace{-10pt}
	\begin{tabular}{l|cc|cc|cc|cc|cc|cc|cc}
		\hline
		\multicolumn{1}{c|}{\textbf{Datasets}} & \multicolumn{2}{c|}{\textbf{Amazon}}                                 & \multicolumn{2}{c|}{\textbf{Yelp}}                                   & \multicolumn{2}{c|}{\textbf{Reddit}}                                 & \multicolumn{2}{c|}{\textbf{Weibo}}                                  & \multicolumn{2}{c|}{\textbf{Facebook}}                               & \multicolumn{2}{c|}{\textbf{Elliptic}}                               & \multicolumn{2}{c}{\textbf{Questions}}                              \\ \hline
		\multicolumn{1}{c|}{\textbf{Methods}}  & \multicolumn{1}{c}{\textbf{AUC}} & \multicolumn{1}{c|}{\textbf{F1}} & \multicolumn{1}{c}{\textbf{AUC}} & \multicolumn{1}{c|}{\textbf{F1}} & \multicolumn{1}{c}{\textbf{AUC}} & \multicolumn{1}{c|}{\textbf{F1}} & \multicolumn{1}{c}{\textbf{AUC}} & \multicolumn{1}{c|}{\textbf{F1}} & \multicolumn{1}{c}{\textbf{AUC}} & \multicolumn{1}{c|}{\textbf{F1}} & \multicolumn{1}{c}{\textbf{AUC}} & \multicolumn{1}{c|}{\textbf{F1}} & \multicolumn{1}{c}{\textbf{AUC}} & \multicolumn{1}{c}{\textbf{F1}} \\ \hline
		MLP                                    & 0.6705                            & 0.4900                           & 0.5999                            & 0.4593                           & 0.4780                            & 0.4920                           & 0.7074                            & 0.4907                           & 0.7247                            & 0.4930                           & 0.7917                            & 0.4944                           & 0.6402                            & 0.4922                           \\
		
		{\color[HTML]{CE6301} ~~+   CLNode}        & {\color[HTML]{CE6301} 0.6275 $\downarrow$} &{\color[HTML]{CE6301} 0.4896} &{\color[HTML]{CE6301} 0.5898 $\downarrow$} & {\color[HTML]{CE6301} 0.4428} & {\color[HTML]{CE6301} 0.6203} & {\color[HTML]{CE6301} 0.4913} & {\color[HTML]{CE6301} 0.8164} & {\color[HTML]{CE6301} 0.4902} & {\color[HTML]{CE6301} 0.7391} & {\color[HTML]{CE6301} 0.4941} & {\color[HTML]{CE6301} 0.6406 $\downarrow$} & {\color[HTML]{CE6301} 0.4945} & {\color[HTML]{CE6301} 0.6093 $\downarrow$} & {\color[HTML]{CE6301} 0.4916} \\
		
		{\color[HTML]{009901} ~~+  RCL}            & {\color[HTML]{009901} 0.6571 $\downarrow$} & {\color[HTML]{009901} 0.4863} & {\color[HTML]{009901}0.6245} & {\color[HTML]{009901} 0.4530} & {\color[HTML]{009901} 0.6565} & {\color[HTML]{009901} 0.4912} & {\color[HTML]{009901} 0.8234} & {\color[HTML]{009901} 0.4899} & {\color[HTML]{009901} 0.7508} & {\color[HTML]{009901} 0.4929} & {\color[HTML]{009901} 0.6812 $\downarrow$} & {\color[HTML]{009901} 0.4942} &{\color[HTML]{009901} 0.6241 $\downarrow$} & {\color[HTML]{009901} 0.4920} \\
		
		{\color[HTML]{3531FF} ~~+  HeteCL}         & {\color[HTML]{3531FF} 0.9890}     & {\color[HTML]{3531FF} 0.4906}    & {\color[HTML]{3531FF} 0.7269}     & {\color[HTML]{3531FF} 0.4609}    & {\color[HTML]{3531FF} 0.6660}     & {\color[HTML]{3531FF} 0.4914}    & {\color[HTML]{3531FF} 0.8838}     & {\color[HTML]{3531FF} 0.4887}    & {\color[HTML]{3531FF} 0.7699}     & {\color[HTML]{3531FF} 0.4953}    & {\color[HTML]{3531FF} 0.8087}     & {\color[HTML]{3531FF} 0.4941}    & {\color[HTML]{3531FF} 0.5997 $\downarrow$}     & {\color[HTML]{3531FF} 0.4812}    \\
		
		{\color[HTML]{6200C9} ~~+  HomoCL}         & {\color[HTML]{6200C9} 0.6995}     & {\color[HTML]{6200C9} 0.4906}    & {\color[HTML]{6200C9} 0.7967}     & {\color[HTML]{6200C9} 0.4609}    & {\color[HTML]{6200C9} 0.7440}     & {\color[HTML]{6200C9} 0.4914}    & {\color[HTML]{6200C9} 0.9741}     & {\color[HTML]{6200C9} 0.4887}    & {\color[HTML]{6200C9} 0.9823}     & {\color[HTML]{6200C9} 0.4953}    & {\color[HTML]{6200C9} 0.9450}     & {\color[HTML]{6200C9} 0.4941}    & {\color[HTML]{6200C9} 0.6726}     & {\color[HTML]{6200C9} 0.4910}    \\
		
		{\color[HTML]{FE0000} ~~+  BCL}            & {\color[HTML]{FE0000} 0.9911}     & {\color[HTML]{FE0000} 0.4906}    & {\color[HTML]{FE0000} 0.7963}     & {\color[HTML]{FE0000} 0.4609}    & {\color[HTML]{FE0000} 0.7486}     & {\color[HTML]{FE0000} 0.4913}    & {\color[HTML]{FE0000} 0.9749}     & {\color[HTML]{FE0000} 0.4887}    & {\color[HTML]{FE0000} 0.9823}     & {\color[HTML]{FE0000} 0.4953}    & {\color[HTML]{FE0000} 0.9427}     & {\color[HTML]{FE0000} 0.4941}    & {\color[HTML]{FE0000} 0.6749}     & {\color[HTML]{FE0000} 0.4929}    \\
		\hline
		GCN                                    & 0.5760                            & 0.4898                           & 0.5608                            & 0.4614                           & 0.5939                            & 0.4913                           & 0.8919                            & 0.5288                           & 0.7977                            & 0.4929                           & 0.5865                            & 0.4952                           & 0.5388                            & 0.4955                           \\
		{\color[HTML]{CE6301} ~~+   CLNode}        & {\color[HTML]{CE6301} 0.8129} &{\color[HTML]{CE6301} 0.4897} &{\color[HTML]{CE6301} 0.5835} & {\color[HTML]{CE6301} 0.4617} & {\color[HTML]{CE6301} 0.6567} & {\color[HTML]{CE6301} 0.4903} & {\color[HTML]{CE6301} 0.9644} & {\color[HTML]{CE6301} 0.4907} & {\color[HTML]{CE6301} 0.9647} & {\color[HTML]{CE6301} 0.4953} & {\color[HTML]{CE6301} 0.6323} & {\color[HTML]{CE6301} 0.4942} & {\color[HTML]{CE6301} 0.6024} & {\color[HTML]{CE6301} 0.4927} \\
		
 		{\color[HTML]{009901} ~~+ RCL}             & {\color[HTML]{009901} 0.8222} & {\color[HTML]{009901} 0.4893} & {\color[HTML]{009901} 0.5848} & {\color[HTML]{009901} 0.4724} & {\color[HTML]{009901} 0.6616} & {\color[HTML]{009901} 0.4908} & {\color[HTML]{009901} 0.9693} & {\color[HTML]{009901} 0.4968} & {\color[HTML]{009901} 0.9711} & {\color[HTML]{009901} 0.4957} & {\color[HTML]{009901} 0.6590} & {\color[HTML]{009901} 0.4983} & {\color[HTML]{009901} 0.6051} & {\color[HTML]{009901} 0.4944} \\ 
		
		{\color[HTML]{3531FF} ~~+  HeteCL}         & {\color[HTML]{3531FF} 0.6678}     & {\color[HTML]{3531FF} 0.4885}    & {\color[HTML]{3531FF} 0.5636}     & {\color[HTML]{3531FF} 0.4433}    & {\color[HTML]{3531FF} 0.6527}     & {\color[HTML]{3531FF} 0.4912}    & {\color[HTML]{3531FF} 0.9726}     & {\color[HTML]{3531FF} 0.4672}    & {\color[HTML]{3531FF} 0.9020}     & {\color[HTML]{3531FF} 0.4721}    & {\color[HTML]{3531FF} 0.6455}     & {\color[HTML]{3531FF} 0.3521}    & {\color[HTML]{3531FF} 0.6005}     & {\color[HTML]{3531FF} 0.5062}    \\
		
		{\color[HTML]{6200C9} ~~+  HomoCL}         & {\color[HTML]{6200C9} 0.8792}     & {\color[HTML]{6200C9} 0.4885}    & {\color[HTML]{6200C9} 0.5988}     & {\color[HTML]{6200C9} 0.4593}    & {\color[HTML]{6200C9} 0.6625}     & {\color[HTML]{6200C9} 0.4917}    & {\color[HTML]{6200C9} 0.9509}     & {\color[HTML]{6200C9} 0.4901}    & {\color[HTML]{6200C9} 0.9984}     & {\color[HTML]{6200C9} 0.4892}    & {\color[HTML]{6200C9} 0.8946}     & {\color[HTML]{6200C9} 0.4942}    & {\color[HTML]{6200C9} 0.5925}     & {\color[HTML]{6200C9} 0.4920}    \\
		
		{\color[HTML]{FE0000} ~~+  BCL}            & {\color[HTML]{FE0000} 0.8316}     & {\color[HTML]{FE0000} 0.4885}    & {\color[HTML]{FE0000} 0.5972}     & {\color[HTML]{FE0000} 0.4536}    & {\color[HTML]{FE0000} 0.6708}     & {\color[HTML]{FE0000} 0.4917}    & {\color[HTML]{FE0000} 0.9742}     & {\color[HTML]{FE0000} 0.4978}    & {\color[HTML]{FE0000} 0.9992}     & {\color[HTML]{FE0000} 0.4930}    & {\color[HTML]{FE0000} 0.8884}     & {\color[HTML]{FE0000} 0.4942}    & {\color[HTML]{FE0000} 0.6142}     & {\color[HTML]{FE0000} 0.4950}    \\
		\hline
		GAT                                    & 0.6455                            & 0.4906                           & 0.5131                            & 0.4613                           & 0.4903                            & 0.4908                           & 0.6309                            & 0.4904                           & 0.9794                            & 0.4929                           & 0.5558                            & 0.4945                           & 0.5169                            & 0.4927                           \\
		
		{\color[HTML]{CE6301} ~~+   CLNode}        & {\color[HTML]{CE6301} 0.6705} &{\color[HTML]{CE6301} 0.4888} &{\color[HTML]{CE6301} 0.5388} & {\color[HTML]{CE6301} 0.4605} & {\color[HTML]{CE6301} 0.6553} & {\color[HTML]{CE6301} 0.4915} & {\color[HTML]{CE6301} 0.8666} & {\color[HTML]{CE6301} 0.4887} & {\color[HTML]{CE6301} 0.9706} & {\color[HTML]{CE6301} 0.4894} & {\color[HTML]{CE6301} 0.6417} & {\color[HTML]{CE6301} 0.9651} & {\color[HTML]{CE6301} 0.5897} & {\color[HTML]{CE6301} 0.5521} \\
		
		{\color[HTML]{009901} ~~+ RCL}             & {\color[HTML]{009901} 0.7060} & {\color[HTML]{009901} 0.4890} & {\color[HTML]{009901} 0.5502} & {\color[HTML]{009901} 0.4571} & {\color[HTML]{009901} 0.6615} & {\color[HTML]{009901} 0.4912} & {\color[HTML]{009901} 0.9048} & {\color[HTML]{009901} 0.4903} & {\color[HTML]{009901} 0.9809} & {\color[HTML]{009901} 0.4926} & {\color[HTML]{009901} 0.6745} & {\color[HTML]{009901} 0.4780} & {\color[HTML]{009901} 0.6093} & {\color[HTML]{009901} 0.4649} \\ 
		
		{\color[HTML]{3531FF} ~~+  HeteCL}         & {\color[HTML]{3531FF} 0.8739}     & {\color[HTML]{3531FF} 0.4906}    & {\color[HTML]{3531FF} 0.5500}     & {\color[HTML]{3531FF} 0.4611}    & {\color[HTML]{3531FF} 0.6707}     & {\color[HTML]{3531FF} 0.4920}    & {\color[HTML]{3531FF} 0.9391}     & {\color[HTML]{3531FF} 0.4901}    & {\color[HTML]{3531FF} 0.9976}     & {\color[HTML]{3531FF} 0.4976}    & {\color[HTML]{3531FF} 0.6296}     & {\color[HTML]{3531FF} 0.4943}    & {\color[HTML]{3531FF} 0.5912}     & {\color[HTML]{3531FF} 0.4920}    \\
		
		{\color[HTML]{6200C9} ~~+  HomoCL}         & {\color[HTML]{6200C9} 0.6506}     & {\color[HTML]{6200C9} 0.4906}    & {\color[HTML]{6200C9} 0.5912}     & {\color[HTML]{6200C9} 0.4611}    & {\color[HTML]{6200C9} 0.6547}     & {\color[HTML]{6200C9} 0.4920}    & {\color[HTML]{6200C9} 0.9448}     & {\color[HTML]{6200C9} 0.4901}    & {\color[HTML]{6200C9} 0.9906}     & {\color[HTML]{6200C9} 0.4976}    & {\color[HTML]{6200C9} 0.8793}     & {\color[HTML]{6200C9} 0.4943}    & {\color[HTML]{6200C9} 0.6521}     & {\color[HTML]{6200C9} 0.4920}    \\
		
		{\color[HTML]{FE0000} ~~+  BCL}            & {\color[HTML]{FE0000} 0.8844}     & {\color[HTML]{FE0000} 0.4906}    & {\color[HTML]{FE0000} 0.5923}     & {\color[HTML]{FE0000} 0.4611}    & {\color[HTML]{FE0000} 0.6845}     & {\color[HTML]{FE0000} 0.4920}    & {\color[HTML]{FE0000} 0.9537}     & {\color[HTML]{FE0000} 0.4901}    & {\color[HTML]{FE0000} 0.9860}     & {\color[HTML]{FE0000} 0.4976}    & {\color[HTML]{FE0000} 0.8768}     & {\color[HTML]{FE0000} 0.4943}    & {\color[HTML]{FE0000} 0.6519}     & {\color[HTML]{FE0000} 0.4920}    \\
		\hline
		GraphSAGE                              & 0.6506                            & 0.4903                           & 0.7067                            & 0.4607                           & 0.5225                            & 0.4909                           & 0.5954                            & 0.4909                           & 0.8988                            & 0.4929                           & 0.8486                            & 0.4944                           & 0.5090                            & 0.4925                           \\
		{\color[HTML]{CE6301} ~~+   CLNode}        & {\color[HTML]{CE6301} 0.7579} &{\color[HTML]{CE6301} 0.4901} &{\color[HTML]{CE6301} 0.6673} & {\color[HTML]{CE6301} 0.4598} & {\color[HTML]{CE6301} 0.6320} & {\color[HTML]{CE6301} 0.4913} & {\color[HTML]{CE6301} 0.7999} & {\color[HTML]{CE6301} 0.4873} & {\color[HTML]{CE6301} 0.9025} & {\color[HTML]{CE6301} 0.4953} & {\color[HTML]{CE6301} 0.6856 $\downarrow$} & {\color[HTML]{CE6301} 0.4943} & {\color[HTML]{CE6301} 0.5760} & {\color[HTML]{CE6301} 0.4586} \\
		
		{\color[HTML]{009901} ~~+ RCL}             & {\color[HTML]{009901} 0.7864} & {\color[HTML]{009901} 0.4898} & {\color[HTML]{009901} 0.6982} & {\color[HTML]{009901} 0.4567} & {\color[HTML]{009901} 0.6837} & {\color[HTML]{009901} 0.4925} & {\color[HTML]{009901} 0.8419} & {\color[HTML]{009901} 0.4880} & {\color[HTML]{009901} 0.9101} & {\color[HTML]{009901} 0.4981} & {\color[HTML]{009901} 0.7005 $\downarrow$} & {\color[HTML]{009901} 0.4946} & {\color[HTML]{009901} 0.5875} & {\color[HTML]{009901} 0.4668} \\ 
		
		{\color[HTML]{3531FF} ~~+  HeteCL}         & {\color[HTML]{3531FF} 0.7474}     & {\color[HTML]{3531FF} 0.4889}    & {\color[HTML]{3531FF} 0.7017}     & {\color[HTML]{3531FF} 0.4594}    & {\color[HTML]{3531FF} 0.6160}     & {\color[HTML]{3531FF} 0.4930}    & {\color[HTML]{3531FF} 0.9636}     & {\color[HTML]{3531FF} 0.4884}    & {\color[HTML]{3531FF} 0.9500}     & {\color[HTML]{3531FF} 0.4941}    & {\color[HTML]{3531FF} 0.8382 $\downarrow$}     & {\color[HTML]{3531FF} 0.4945}    & {\color[HTML]{3531FF} 0.6142}     & {\color[HTML]{3531FF} 0.4932}    \\
		
		{\color[HTML]{6200C9} ~~+  HomoCL}         & {\color[HTML]{6200C9} 0.9730}     & {\color[HTML]{6200C9} 0.4889}    & {\color[HTML]{6200C9} 0.8167}     & {\color[HTML]{6200C9} 0.4594}    & {\color[HTML]{6200C9} 0.7425}     & {\color[HTML]{6200C9} 0.4930}    & {\color[HTML]{6200C9} 0.9870}     & {\color[HTML]{6200C9} 0.4884}    & {\color[HTML]{6200C9} 1.0000}     & {\color[HTML]{6200C9} 1.0000}    & {\color[HTML]{6200C9} 0.9496}     & {\color[HTML]{6200C9} 0.4940}    & {\color[HTML]{6200C9} 0.6936}     & {\color[HTML]{6200C9} 0.4932}    \\
		
		{\color[HTML]{FE0000} ~~+  BCL}            & {\color[HTML]{FE0000} 0.9925}     & {\color[HTML]{FE0000} 0.4889}    & {\color[HTML]{FE0000} 0.8162}     & {\color[HTML]{FE0000} 0.4594}    & {\color[HTML]{FE0000} 0.7397}     & {\color[HTML]{FE0000} 0.4930}    & {\color[HTML]{FE0000} 0.9873}     & {\color[HTML]{FE0000} 0.4884}    & {\color[HTML]{FE0000} 1.0000}     & {\color[HTML]{FE0000} 1.0000}    & {\color[HTML]{FE0000} 0.9481}     & {\color[HTML]{FE0000} 0.4945}    & {\color[HTML]{FE0000} 0.6951}     & {\color[HTML]{FE0000} 0.4932}    \\ 
		\hline
		CARE-GNN                               & 0.9244                            & 0.8492                           & 0.6949                            & 0.5695                           & 0.6012                            & 0.4957                           & 0.8012                            & 0.7035                           & 0.7269                            & 0.4674                           & 0.6743                            & 0.4807                           & 0.5873                            & 0.5087                           \\
		
		{\color[HTML]{CE6301} ~~+ CLNode}          & {\color[HTML]{CE6301} 0.9312} & {\color[HTML]{CE6301} 0.8598} & {\color[HTML]{CE6301} 0.6853 $\downarrow$} & {\color[HTML]{CE6301} 0.5557} & {\color[HTML]{CE6301} 0.6028} & {\color[HTML]{CE6301} 0.4983} & {\color[HTML]{CE6301} 0.8573} & {\color[HTML]{CE6301} 0.7436} & {\color[HTML]{CE6301} 0.7316} & {\color[HTML]{CE6301} 0.5034} & {\color[HTML]{CE6301} 0.6712 $\downarrow$} & {\color[HTML]{CE6301} 0.4812} & {\color[HTML]{CE6301} 0.5891} & {\color[HTML]{CE6301} 0.5317} \\ 
		
		{\color[HTML]{009901} ~~+ RCL}             & {\color[HTML]{009901} 0.9388} & {\color[HTML]{009901} 0.8594} & {\color[HTML]{009901} 0.6953} & {\color[HTML]{009901} 0.5674} & {\color[HTML]{009901} 0.6032} & {\color[HTML]{009901} 0.4985} & {\color[HTML]{009901} 0.8720} & {\color[HTML]{009901} 0.7639} & {\color[HTML]{009901} 0.7433} & {\color[HTML]{009901} 0.5247} & {\color[HTML]{009901} 0.6722 $\downarrow$} & {\color[HTML]{009901} 0.4813} & {\color[HTML]{009901} 0.5960} & {\color[HTML]{009901} 0.5335} \\ 
		
		{\color[HTML]{3531FF} ~~+  HeteCL}         & {\color[HTML]{3531FF} 0.9311}     & {\color[HTML]{3531FF} 0.8545}    & {\color[HTML]{3531FF} 0.7095}     & {\color[HTML]{3531FF} 0.5781}    & {\color[HTML]{3531FF} 0.6097}     & {\color[HTML]{3531FF} 0.4914}    & {\color[HTML]{3531FF} 0.9541}     & {\color[HTML]{3531FF} 0.8213}    & {\color[HTML]{3531FF} 0.7317}     & {\color[HTML]{3531FF} 0.4586}    & {\color[HTML]{3531FF} 0.6967}     & {\color[HTML]{3531FF} 0.4017}    & {\color[HTML]{3531FF} 0.5909}     & {\color[HTML]{3531FF} 0.4764}    \\
		
		{\color[HTML]{6200C9} ~~+  HomoCL}         & {\color[HTML]{6200C9} 0.9383}     & {\color[HTML]{6200C9} 0.8735}    & {\color[HTML]{6200C9} 0.7099}     & {\color[HTML]{6200C9} 0.5999}    & {\color[HTML]{6200C9} 0.6018}     & {\color[HTML]{6200C9} 0.4914}    & {\color[HTML]{6200C9} 0.9485}     & {\color[HTML]{6200C9} 0.8173}    & {\color[HTML]{6200C9} 0.7391}     & {\color[HTML]{6200C9} 0.5283}    & {\color[HTML]{6200C9} 0.6828}     & {\color[HTML]{6200C9} 0.4423}    & {\color[HTML]{6200C9} 0.5930}     & {\color[HTML]{6200C9} 0.5126}    \\
		
		{\color[HTML]{FE0000} ~~+  BCL}            & {\color[HTML]{FE0000} 0.9758}     & {\color[HTML]{FE0000} 0.8804}    & {\color[HTML]{FE0000} 0.7104}     & {\color[HTML]{FE0000} 0.5879}    & {\color[HTML]{FE0000} 0.6130}     & {\color[HTML]{FE0000} 0.5218}    & {\color[HTML]{FE0000} 0.9588}     & {\color[HTML]{FE0000} 0.8560}    & {\color[HTML]{FE0000} 0.7522}     & {\color[HTML]{FE0000} 0.5486}    & {\color[HTML]{FE0000} 0.7158}     & {\color[HTML]{FE0000} 0.5065}    & {\color[HTML]{FE0000} 0.6033}     & {\color[HTML]{FE0000} 0.5348}    \\
		\hline
		PCGNN                                  & 0.9665                            & 0.8823                           & 0.8168                            & 0.6344                           & 0.6258                            & 0.4915                           & 0.8168                            & 0.6344                           & 0.7352                            & 0.5644                           & 0.9204                            & 0.5160                           & 0.6285                            & 0.2634                           \\
		
		{\color[HTML]{CE6301} ~~+ CLNode}          & {\color[HTML]{CE6301} 0.9689} & {\color[HTML]{CE6301} 0.8214} & {\color[HTML]{CE6301} 0.8136 $\downarrow$} & {\color[HTML]{CE6301} 0.6647} & {\color[HTML]{CE6301} 0.6300} & {\color[HTML]{CE6301} 0.4875} & {\color[HTML]{CE6301} 0.9513} & {\color[HTML]{CE6301} 0.8686} & {\color[HTML]{CE6301} 0.7183 $\downarrow$} & {\color[HTML]{CE6301} 0.4924} & {\color[HTML]{CE6301} 0.9000 $\downarrow$} & {\color[HTML]{CE6301} 0.5031} & {\color[HTML]{CE6301} 0.6292} & {\color[HTML]{CE6301} 0.2403} \\ 
		{\color[HTML]{009901} ~~+ RCL}             & {\color[HTML]{009901} 0.9693} & {\color[HTML]{009901} 0.8760} & {\color[HTML]{009901} 0.8146 $\downarrow$} & {\color[HTML]{009901} 0.6502} & {\color[HTML]{009901} 0.6499} & {\color[HTML]{009901} 0.4902} & {\color[HTML]{009901} 0.9520} & {\color[HTML]{009901} 0.8702} & {\color[HTML]{009901} 0.7382} & {\color[HTML]{009901} 0.5724} & {\color[HTML]{009901} 0.9066 $\downarrow$} & {\color[HTML]{009901} 0.5213} & {\color[HTML]{009901} 0.6279 $\downarrow$} & {\color[HTML]{009901} 0.2548} \\
		
		{\color[HTML]{3531FF} ~~+  HeteCL}         & {\color[HTML]{3531FF} 0.9671}     & {\color[HTML]{3531FF} 0.8811}    & {\color[HTML]{3531FF} 0.8182}     & {\color[HTML]{3531FF} 0.6496}    & {\color[HTML]{3531FF} 0.6287}     & {\color[HTML]{3531FF} 0.4915}    & {\color[HTML]{3531FF} 0.9623}     & {\color[HTML]{3531FF} 0.8763}    & {\color[HTML]{3531FF} 0.7452}     & {\color[HTML]{3531FF} 0.6370}    & {\color[HTML]{3531FF} 0.9308}     & {\color[HTML]{3531FF} 0.5525}    & {\color[HTML]{3531FF} 0.6376}     & {\color[HTML]{3531FF} 0.2222}    \\
		
		{\color[HTML]{6200C9} ~~+  HomoCL}         & {\color[HTML]{6200C9} 0.9689}     & {\color[HTML]{6200C9} 0.8842}    & {\color[HTML]{6200C9} 0.8192}     & {\color[HTML]{6200C9} 0.6645}    & {\color[HTML]{6200C9} 0.6645}     & {\color[HTML]{6200C9} 0.4915}    & {\color[HTML]{6200C9} 0.9423}     & 	{\color[HTML]{6200C9} 0.8325}    & {\color[HTML]{6200C9} 0.7538}     & {\color[HTML]{6200C9} 0.4935}    & {\color[HTML]{6200C9} 0.9226}     & {\color[HTML]{6200C9} 0.5172}    & {\color[HTML]{6200C9} 0.6382}     & {\color[HTML]{6200C9} 0.2789}    \\
		
		{\color[HTML]{FE0000} ~~+  BCL}            & {\color[HTML]{FE0000} 0.9686}     & {\color[HTML]{FE0000} 0.8837}    & {\color[HTML]{FE0000} 0.8192}     & {\color[HTML]{FE0000} 0.6712}    & {\color[HTML]{FE0000} 0.6610}     & {\color[HTML]{FE0000} 0.4915}    & {\color[HTML]{FE0000} 0.9646}     & {\color[HTML]{FE0000} 0.8812}    & {\color[HTML]{FE0000} 0.7578}     & {\color[HTML]{FE0000} 0.6226}    & {\color[HTML]{FE0000} 0.9300}     & {\color[HTML]{FE0000} 0.5510}    & {\color[HTML]{FE0000} 0.6382}     & {\color[HTML]{FE0000} 0.2732}    \\
		\hline
		BWGNN                                  & 0.9752                            & 0.9230                           & 0.8217                            & 0.6873                           & 0.6689                           & 0.5333                           & 0.9627                            & 0.9134                           & 0.8115                            & 0.5364                           & 0.8575                            & 0.6876                           & 0.6499                            & 0.5864                           \\
		
		{\color[HTML]{CE6301} ~~+   CLNode}        & {\color[HTML]{CE6301} 0.9829} &{\color[HTML]{CE6301} 0.9295} &{\color[HTML]{CE6301} 0.8280} & {\color[HTML]{CE6301} 0.6927} & {\color[HTML]{CE6301} 0.6522 $\downarrow$} & {\color[HTML]{CE6301} 0.5338} & {\color[HTML]{CE6301} 0.9640} & {\color[HTML]{CE6301} 0.9274} & {\color[HTML]{CE6301} 0.6285 $\downarrow$} & {\color[HTML]{CE6301} 0.5474} & {\color[HTML]{CE6301} 0.9526} & {\color[HTML]{CE6301} 0.7919} & {\color[HTML]{CE6301} 0.6546} & {\color[HTML]{CE6301} 0.5954} \\
		
		{\color[HTML]{009901} ~~+ RCL}             & {\color[HTML]{009901} 0.9845} & {\color[HTML]{009901} 0.9301} & {\color[HTML]{009901} 0.8318} & {\color[HTML]{009901} 0.7007} & {\color[HTML]{009901} 0.6757} & {\color[HTML]{009901} 0.5334} & {\color[HTML]{009901} 0.9667} & {\color[HTML]{009901} 0.9303} & {\color[HTML]{009901} 0.8525} & {\color[HTML]{009901} 0.5695} & {\color[HTML]{009901} 0.9671} & {\color[HTML]{009901} 0.8030} & {\color[HTML]{009901} 0.6661} & {\color[HTML]{009901} 0.6041} \\ 
		
		{\color[HTML]{3531FF} ~~+  HeteCL}         & {\color[HTML]{3531FF} 0.9890}     & {\color[HTML]{3531FF} 0.9298}    & {\color[HTML]{3531FF} 0.8355}     & {\color[HTML]{3531FF} 0.6995}    & {\color[HTML]{3531FF} 0.6712}     & {\color[HTML]{3531FF} 0.5426}    & {\color[HTML]{3531FF} 0.9986}     & {\color[HTML]{3531FF} 0.9726}    & {\color[HTML]{3531FF} 1.0000}     & {\color[HTML]{3531FF} 1.0000}    & {\color[HTML]{3531FF} 0.9890}     & {\color[HTML]{3531FF} 0.8240}    & {\color[HTML]{3531FF} 0.9625}     & {\color[HTML]{3531FF} 0.7094}    \\
		
		{\color[HTML]{6200C9} ~~+  HomoCL}         & {\color[HTML]{6200C9} 0.9848}     & {\color[HTML]{6200C9} 0.9340}    & {\color[HTML]{6200C9} 0.8459}     & {\color[HTML]{6200C9} 0.7108}    & {\color[HTML]{6200C9} 0.6752}     & {\color[HTML]{6200C9} 0.5375}    & {\color[HTML]{6200C9} 0.9997}     & {\color[HTML]{6200C9} 0.9872}    & {\color[HTML]{6200C9} 1.0000}     & {\color[HTML]{6200C9} 1.0000}    & {\color[HTML]{6200C9} 0.9901}     & {\color[HTML]{6200C9} 0.8262}    & {\color[HTML]{6200C9} 0.9572}     & {\color[HTML]{6200C9} 0.7047}    \\
		
		{\color[HTML]{FE0000} ~~+  BCL}            & {\color[HTML]{FE0000} 0.9923}     & {\color[HTML]{FE0000} 0.9307}    & {\color[HTML]{FE0000} 0.8489}     & {\color[HTML]{FE0000} 0.7241}    & {\color[HTML]{FE0000} 0.6826}     & {\color[HTML]{FE0000} 0.5336}    & {\color[HTML]{FE0000} 0.9997}     & {\color[HTML]{FE0000} 0.9872}    & {\color[HTML]{FE0000} 1.0000}     & {\color[HTML]{FE0000} 1.0000}    & {\color[HTML]{FE0000} 0.9914}     & {\color[HTML]{FE0000} 0.8289}    & {\color[HTML]{FE0000} 0.9729}     & {\color[HTML]{FE0000} 0.7281}    \\
		\hline
		AMNET                                  & 0.9167                            & 0.6558                           & 0.8358                            & 0.5092                           & 0.6677                            & 0.4827                           & 0.9179                            & 0.3810                           & 0.7492                            & 0.2500                           & 0.8773                            & 0.5074                           & 0.6174                            & 0.5873                           \\
		
		{\color[HTML]{CE6301} ~~+ CLNode}          & {\color[HTML]{CE6301} 0.9228} & {\color[HTML]{CE6301} 0.6565} & {\color[HTML]{CE6301} 0.8498} & {\color[HTML]{CE6301} 0.5274} & {\color[HTML]{CE6301} 0.7272} & {\color[HTML]{CE6301} 0.5111} & {\color[HTML]{CE6301} 0.9375} & {\color[HTML]{CE6301} 0.5018} & {\color[HTML]{CE6301} 0.8317} & {\color[HTML]{CE6301} 0.6581} & {\color[HTML]{CE6301} 0.9164} & {\color[HTML]{CE6301} 0.6885} & {\color[HTML]{CE6301} 0.7412} & {\color[HTML]{CE6301} 0.6988} \\
		
		{\color[HTML]{009901} ~~+ RCL}             & {\color[HTML]{009901} 0.9302} & {\color[HTML]{009901} 0.6655} & {\color[HTML]{009901} 0.8594} & {\color[HTML]{009901} 0.5286} & {\color[HTML]{009901} 0.7476} & {\color[HTML]{009901} 0.5195} & {\color[HTML]{009901} 0.9426} & {\color[HTML]{009901} 0.5376} & {\color[HTML]{009901} 0.8512} & {\color[HTML]{009901} 0.6831} & {\color[HTML]{009901} 0.9345} & {\color[HTML]{009901} 0.7010} & {\color[HTML]{009901} 0.8412} & {\color[HTML]{009901} 0.7926} \\
		
		{\color[HTML]{3531FF} ~~+  HeteCL}         & {\color[HTML]{3531FF} 0.9211}     & {\color[HTML]{3531FF} 0.6591}    & {\color[HTML]{3531FF} 0.8795}     & {\color[HTML]{3531FF} 0.5168}    & {\color[HTML]{3531FF} 0.7622}     & {\color[HTML]{3531FF} 0.4999}    & {\color[HTML]{3531FF} 0.9862}     & {\color[HTML]{3531FF} 0.6210}    & {\color[HTML]{3531FF} 1.0000}     & {\color[HTML]{3531FF} 1.0000}    & {\color[HTML]{3531FF} 0.9611}     & {\color[HTML]{3531FF} 0.7888}    & {\color[HTML]{3531FF} 0.9960}     & {\color[HTML]{3531FF} 0.9443}    \\
		
		{\color[HTML]{6200C9} ~~+  HomoCL}         & {\color[HTML]{6200C9} 0.9349}     & {\color[HTML]{6200C9} 0.6633}    & {\color[HTML]{6200C9} 0.8808}     & {\color[HTML]{6200C9} 0.5358}    & {\color[HTML]{6200C9} 0.7716}     & {\color[HTML]{6200C9} 0.5076}    & {\color[HTML]{6200C9} 0.9879}     & {\color[HTML]{6200C9} 0.6396}    & {\color[HTML]{6200C9} 1.0000}     & {\color[HTML]{6200C9} 1.0000}    & {\color[HTML]{6200C9} 0.9670}     & {\color[HTML]{6200C9} 0.7940}    & {\color[HTML]{6200C9} 0.9895}     & {\color[HTML]{6200C9} 0.9427}    \\
		
		{\color[HTML]{FE0000} ~~+  BCL}            & {\color[HTML]{FE0000} 0.9377}     & {\color[HTML]{FE0000} 0.6691}    & {\color[HTML]{FE0000} 0.8897}     & {\color[HTML]{FE0000} 0.5463}    & {\color[HTML]{FE0000} 0.7911}     & {\color[HTML]{FE0000} 0.5329}    & {\color[HTML]{FE0000} 0.9919}     & {\color[HTML]{FE0000} 0.6444}    & {\color[HTML]{FE0000} 1.0000}     & {\color[HTML]{FE0000} 1.0000}    & {\color[HTML]{FE0000} 0.9743}     & {\color[HTML]{FE0000} 0.7996}    & {\color[HTML]{FE0000} 0.9972}     & {\color[HTML]{FE0000} 0.9571}    \\
		\hline
		
		SplitGNN                               & 0.9193                            & 0.7090                           & 0.9184                            & 0.7525                           & 0.6548                            & 0.4116                           & 0.9132                            & 0.8313                           & 0.8510                            & 0.4941                           & 0.9364                            & 0.5925                           & 0.6163                            & 0.5841                           \\
		
		{\color[HTML]{CE6301} ~~+ CLNode}          & {\color[HTML]{CE6301} 0.9084 $\downarrow$} & {\color[HTML]{CE6301} 0.7062} & {\color[HTML]{CE6301} 0.9158 $\downarrow$} & {\color[HTML]{CE6301} 0.7388} & {\color[HTML]{CE6301} 0.6814} & {\color[HTML]{CE6301} 0.4145} & {\color[HTML]{CE6301} 0.9268} & {\color[HTML]{CE6301} 0.8325} & {\color[HTML]{CE6301} 0.8877} & {\color[HTML]{CE6301} 0.4943} & {\color[HTML]{CE6301} 0.9397} & {\color[HTML]{CE6301} 0.6028} & {\color[HTML]{CE6301} 0.6473} & {\color[HTML]{CE6301} 0.5836} \\ 
		
		{\color[HTML]{009901} ~~+ RCL}             & {\color[HTML]{009901} 0.9111 $\downarrow$} & {\color[HTML]{009901} 0.7033} & {\color[HTML]{009901} 0.9176 $\downarrow$} & {\color[HTML]{009901} 0.7435} & {\color[HTML]{009901} 0.6992} & {\color[HTML]{009901} 0.4156} & {\color[HTML]{009901} 0.9299} & {\color[HTML]{009901} 0.8397} & {\color[HTML]{009901} 0.8905} & {\color[HTML]{009901} 0.4921} & {\color[HTML]{009901} 0.9376} & {\color[HTML]{009901} 0.6027} & {\color[HTML]{009901} 0.6434} & {\color[HTML]{009901} 0.5823} \\
		
		{\color[HTML]{3531FF} ~~+  HeteCL}         & {\color[HTML]{3531FF} 0.9324}     & {\color[HTML]{3531FF} 0.7026}    & {\color[HTML]{3531FF} 0.9188}     & {\color[HTML]{3531FF} 0.7102}    & {\color[HTML]{3531FF} 0.7015}     & {\color[HTML]{3531FF} 0.4127}    & {\color[HTML]{3531FF} 0.9411}     & {\color[HTML]{3531FF} 0.6639}    & {\color[HTML]{3531FF} 0.8818}     & {\color[HTML]{3531FF} 0.7822}    & {\color[HTML]{3531FF} 0.9373}     & {\color[HTML]{3531FF} 0.5950}    & {\color[HTML]{3531FF} 0.6563}     & {\color[HTML]{3531FF} 0.4499}    \\
		
		{\color[HTML]{6200C9} ~~+  HomoCL}         & {\color[HTML]{6200C9} 0.9287}     & {\color[HTML]{6200C9} 0.7147}    & {\color[HTML]{6200C9} 0.9210}     & {\color[HTML]{6200C9} 0.7369}    & {\color[HTML]{6200C9} 0.7183}     & {\color[HTML]{6200C9} 0.4263}    & {\color[HTML]{6200C9} 0.9169}     & {\color[HTML]{6200C9} 0.6489}    & {\color[HTML]{6200C9} 0.8344}     & {\color[HTML]{6200C9} 0.4912}    & {\color[HTML]{6200C9} 0.9403}     & {\color[HTML]{6200C9} 0.6009}    & {\color[HTML]{6200C9} 0.6581}     & {\color[HTML]{6200C9} 0.4180}    \\
		
		{\color[HTML]{FE0000} ~~+  BCL}            & {\color[HTML]{FE0000} 0.9334}     & {\color[HTML]{FE0000} 0.7086}    & {\color[HTML]{FE0000} 0.9216}     & {\color[HTML]{FE0000} 0.7273}    & {\color[HTML]{FE0000} 0.7188}     & {\color[HTML]{FE0000} 0.4280}    & {\color[HTML]{FE0000} 0.9393}     & {\color[HTML]{FE0000} 0.6648}    & {\color[HTML]{FE0000} 0.9272}     & {\color[HTML]{FE0000} 0.4935}    & {\color[HTML]{FE0000} 0.9427}     & {\color[HTML]{FE0000} 0.6203}    & {\color[HTML]{FE0000} 0.6801}     & {\color[HTML]{FE0000} 0.4554}    \\
		\hline
		
		GHRN                                   & 0.9771                            & 0.9109                           & 0.8374                            & 0.7126                           & 0.7246                            & 0.5599                           & 0.9686                            & 0.9108                           & 0.8499                            & 0.5604                           & 0.9507                            & 0.7741                           & 0.7403                            & 0.5820                           \\
		
		{\color[HTML]{CE6301} ~~+ CLNode}          & {\color[HTML]{CE6301} 0.9635 $\downarrow$} & {\color[HTML]{CE6301} 0.8997} & {\color[HTML]{CE6301} 0.8374} & {\color[HTML]{CE6301} 0.7200} & {\color[HTML]{CE6301} 0.7374} & {\color[HTML]{CE6301} 0.5687} & {\color[HTML]{CE6301} 0.9722} & {\color[HTML]{CE6301} 0.9281} & {\color[HTML]{CE6301} 0.9524} & {\color[HTML]{CE6301} 0.5812} & {\color[HTML]{CE6301} 0.9558} & {\color[HTML]{CE6301} 0.7986} & {\color[HTML]{CE6301} 0.7823} & {\color[HTML]{CE6301} 0.5578} \\
		
		{\color[HTML]{009901} ~~+ RCL}             & {\color[HTML]{009901} 0.9785} & {\color[HTML]{009901} 0.8966} & {\color[HTML]{009901} 0.8450} & {\color[HTML]{009901} 0.5987} & {\color[HTML]{009901} 0.7428} & {\color[HTML]{009901} 0.5635} & {\color[HTML]{009901} 0.9873} & {\color[HTML]{009901} 0.9589} & {\color[HTML]{009901} 0.9637} & {\color[HTML]{009901} 0.6034} & {\color[HTML]{009901} 0.9619} & {\color[HTML]{009901} 0.8038} & {\color[HTML]{009901} 0.8046} & {\color[HTML]{009901} 0.5845} \\
		
		{\color[HTML]{3531FF} ~~+  HeteCL}         & {\color[HTML]{3531FF} 0.9899}     & {\color[HTML]{3531FF} 0.8964}    & {\color[HTML]{3531FF} 0.8464}     & {\color[HTML]{3531FF} 0.7136}    & {\color[HTML]{3531FF} 0.7462}     & {\color[HTML]{3531FF} 0.5858}    & {\color[HTML]{3531FF} 0.9999}     & {\color[HTML]{3531FF} 0.9825}    & {\color[HTML]{3531FF} 1.0000}     & {\color[HTML]{3531FF} 1.0000}    & {\color[HTML]{3531FF} 0.9921}     & {\color[HTML]{3531FF} 0.8485}    & {\color[HTML]{3531FF} 0.9813}     & {\color[HTML]{3531FF} 0.7864}    \\
		
		{\color[HTML]{6200C9} ~~+  HomoCL}         & {\color[HTML]{6200C9} 0.9831}     & {\color[HTML]{6200C9} 0.9236}    & {\color[HTML]{6200C9} 0.8493}     & {\color[HTML]{6200C9} 0.7184}    & {\color[HTML]{6200C9} 0.7556}     & {\color[HTML]{6200C9} 0.5837}    & {\color[HTML]{6200C9} 0.9999}     & {\color[HTML]{6200C9} 0.9828}    & {\color[HTML]{6200C9} 1.0000}     & {\color[HTML]{6200C9} 1.0000}    & {\color[HTML]{6200C9} 0.9929}     & {\color[HTML]{6200C9} 0.8484}    & {\color[HTML]{6200C9} 0.9819}     & {\color[HTML]{6200C9} 0.7956}    \\
		
		{\color[HTML]{FE0000} ~~+  BCL}            & {\color[HTML]{FE0000} 0.9894}     & {\color[HTML]{FE0000} 0.8604}    & {\color[HTML]{FE0000} 0.8532}     & {\color[HTML]{FE0000} 0.5294}    & {\color[HTML]{FE0000} 0.7547}     & {\color[HTML]{FE0000} 0.5887}    & {\color[HTML]{FE0000} 0.9999}     & {\color[HTML]{FE0000} 0.9897}    & {\color[HTML]{FE0000} 1.0000}     & {\color[HTML]{FE0000} 1.0000}    & {\color[HTML]{FE0000} 0.9936}     & {\color[HTML]{FE0000} 0.8501}    & {\color[HTML]{FE0000} 0.9863}     & {\color[HTML]{FE0000} 0.8022}  \\
		\hline 
	\end{tabular}
	\caption{Improved performance on seven datasets.}
%	\vspace{-10pt}
	\label{Table 2}
\end{table*}

\subsubsection{Implementation Details.}
Additionally, the hyperparameter $\alpha$, which serves to balance the contributions of the two directional models, was tuned across the range \{0.1, 0.2, 0.3, ..., 0.9\}.
$\lambda_0$ is varied in \{0.1, 0.2, ... , 0.9\} and $T$ is varied from 10\% to 90\% of the original method's training rounds.
\subsubsection{Existing Detector.}
To rigorously assess the efficacy of the BCL framework, we compared it against detectors based on MLP\cite{mlp}, GCN\cite{gcn}, GAT\cite{gat}, and GraphSAGE\cite{graphsage}, as well as six state-of-the-art supervised anomaly detectors: PC-GNN\cite{pcgnn}, BWGNN\cite{bwgnn}, CARE-GNN\cite{caregnn}, AMNet\cite{amnet}, GHRN\cite{ghrn}, and Split-GNN\cite{splitgnn}.
\subsubsection{Experimental Details.}
Avatar
In this study, all existing detectors were trained in strict accordance with their original specifications. The dataset was partitioned into training (40\%), testing (40\%), and validation (20\%) sets. Performance was evaluated using Macro-F1 and AUC metrics, along with their percentage improvements. All experimental validations were conducted on a high-performance server equipped with a 12-core CPU and a single Nvidia 3090Ti GPU, featuring 32GB of dedicated RAM.
\subsection{Improvement on Anomaly Detection Performance (RQ1 and RQ2)}
The results in Table \ref{Table 2} show that the introduction of CLNode and RCL on the basis of the existing GAD method has a relatively good improvement on most of the detection performance. This shows that by prioritizing the learning of simpler nodes, the GAD model can better access the feature information of the graph, thus improving the ability to identify abnormal nodes. However, on some datasets, the performance improvement is not significant. For example, having CLNode combined with GCN applied on the Amazon dataset, the detection even showed a decrease. This suggests that the graph course learning approach may not be fully adapted to the GAD task, further highlighting the need to propose a new framework.

As can be seen from Table \ref{Table 2}, BCL outperforms existing graph course learning methods on all datasets, and on some datasets, e.g., Elliptic, GCN is able to improve its performance by almost 30$\%$ after applying BCL. In particular, GraphSAGE, BWGNN, AMNET, and GHRN combined with BCL achieve AUC and F1 scores of 1 on the Facebook dataset, indicating that these models are able to perfectly distinguish between anomalous and non-anomalous nodes. To the best of our knowledge, this is something that no previous study has been able to achieve. This demonstrates the effectiveness of our proposed training strategy.The BCL framework provides a more comprehensive view by considering both homogeneity and heterogeneity, which may be the key to its performance improvement.
\subsection{Ablation Study (RQ3)}
In the ablation study, HomoCL represents the GAD model focusing on homogeneity, prioritizing nodes with lower bi-directional difficulty scores for training, while HeteCL focuses on heterogeneity. As shown in Table 2, the individual performances of HomoCL and HeteCL vary significantly across different datasets, with complementary effects observed in some cases but similar performances in others. This highlights the limitations of relying solely on one direction (either homogeneity or heterogeneity). For instance, on the Amazon dataset, the combination of HomoCL and GAT fails to deliver satisfactory results, whereas HeteCL excels. This further underscores the necessity of considering both homogeneity and heterogeneity in anomaly detection. Our proposed BCL strategy is precisely based on the fact that the GAD model can handle both homogeneity and heterogeneity, providing more robust anomaly detection results by integrating the advantages of both directions.

\subsection{The Impact of Training Scheduler (RQ4)} 

\begin{table}[h]
	\centering
	\setlength{\tabcolsep}{7pt}
	\scriptsize
	\begin{tabular}{c|c|c|c|c|c}
		\hline
		\multirow{2}{*}{Dataset} & \multicolumn{1}{c|}{\multirow{2}{*}{Method}} & \multicolumn{4}{c}{Pacing Function}                                       \\ \cline{3-6} 
		& \multicolumn{1}{c|}{}                        & \multicolumn{1}{c|}{linear} & \multicolumn{1}{c|}{root} & geomo        & \multicolumn{1}{c}{none}    \\ \hline
		\multirow{10}{*}{Amazon} & MLP + BCL                                    & 0.9892                      & \textbf{0.9911}           & 0.9835   & 0.8512       \\
		& GCN + BCL                                    & \textbf{0.8316}             & 0.8048                    & 0.7794        & 0.7052  \\
		& GAT + BCL                                    & \textbf{0.8844}             & 0.8572                    & 0.8342        & 0.8280  \\
		& GraphSAGE + BCL                              & 0.9729                      & \textbf{0.9925}           & 0.9915        & 0.8496  \\
		& CARE-GNN + BCL                               & 0.9437                      & 0.9656                    & \textbf{0.9758} &0.9364 \\
		& PCGNN+BCL                                    & \textbf{0.9686}             & 0.9656                    & 0.9657        & 0.9611 \\
		& BWGNN + BCL                                  & \textbf{0.9923}             & 0.9869                    & 0.9878        & 0.9789  \\
		& AMNET + BCL                                  & \textbf{0.9377}             & 0.9328                    & 0.9284        & 0.9227  \\
		& SplitGNN + BCL                               & \textbf{0.9334}             & 0.9273                    & 0.9246        & 0.9149  \\
		& GHRN + BCL                                   & \textbf{0.9894}             & 0.9858                    & 0.9864        & 0.9618  \\
		\hline
	\end{tabular}
	\caption{Comparisons between different pacing functions}
	\label{Table 3}
\end{table}
We evaluated in detail the sensitivity of BCL on three different pacing functions: linear, rooted and geometric. In Table \ref{Table 3}, by comparing the performance of different pacing functions on the Amazon dataset, we find that the linear pacing function shows a more obvious advantage on most methods. This indicates that the linear pacing function can effectively guide the model learning by gradually increasing the sample difficulty, so that the model can quickly master the basic features in the early stage of training, and then smoothly transition to the complex samples. This smooth transition helps the model to establish a deep understanding of the data, improving the learning efficiency and final performance. In contrast, the rootedness pacing function introduces difficult samples quickly at the beginning, but may lead to unstable learning. The geometric pacing function, while performing well in some cases, is not as stable as the linear pacing function. A reasonable training scheduling strategy ensures that the model is gradually exposed to samples of increasing difficulty, improving learning efficiency and performance. Therefore, choosing an appropriate pacing function is crucial for the application of the BCL framework to the GAD task, and the linear pacing function is preferred due to its stability and effectiveness.
\subsection{The Impact of Balancing Parameter $\alpha$ (RQ5)}
\begin{figure}[htbp]
	\centering
	\includegraphics[width=1\linewidth]{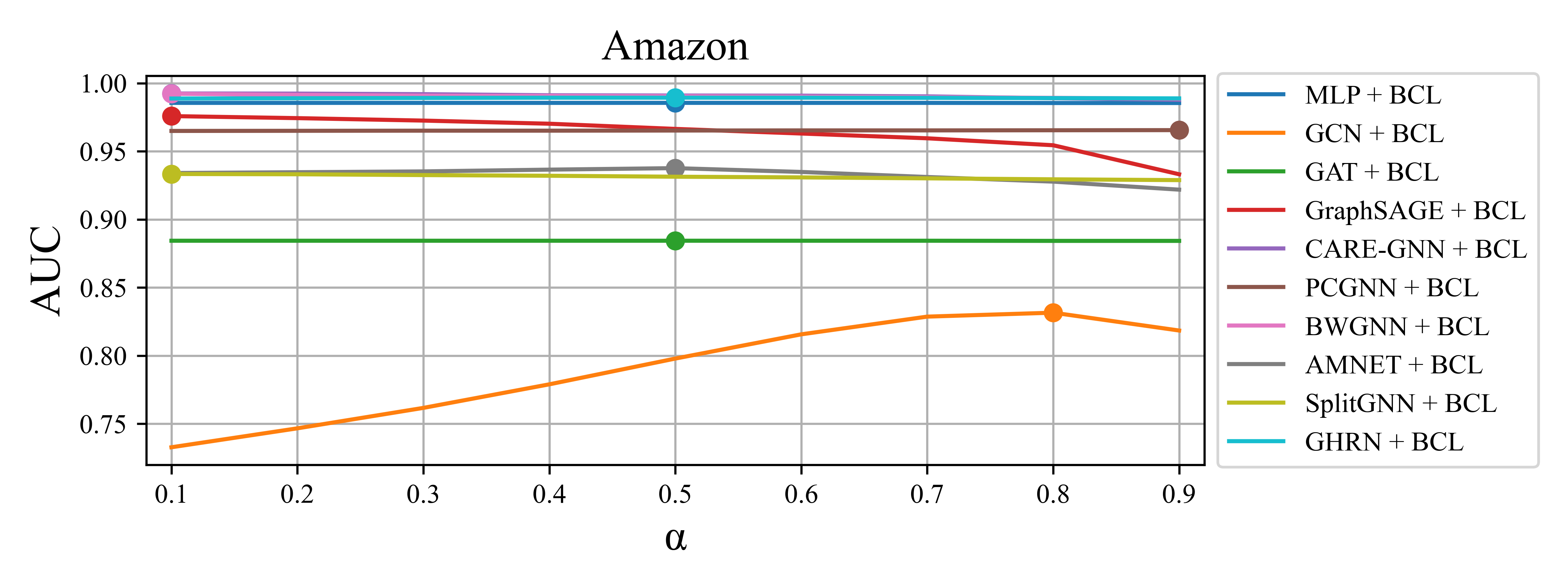} 
	\vspace{-10pt}
	\caption{The Impact of different $\alpha$ on the performance}
	\label{fig:4}
\end{figure}
The parameter $\alpha$ serves to balance the BCL's focus between homogeneity and heterogeneity. Specifically, a larger $\alpha$ directs the framework to emphasize homogeneous features, whereas a smaller $\alpha$ shifts the focus towards heterogeneous features. As depicted in Figure \ref{fig:3}, on the Amazon dataset, GAD such as MLP, GAT, AMNET, and GHRN achieve optimal performance at $\alpha=0.5$, indicating a well-balanced consideration of both homogeneity and heterogeneity. Conversely, GraphSAGE, CARE-GNN, BWGNN, and Split-GNN exhibit peak performance at $\alpha$=0.1, suggesting a heightened sensitivity to heterogeneous features. Notably, GCN and PC-GNN perform best at $\alpha$=0.8 and $\alpha$=0.9, respectively, which implies a greater responsiveness to homogeneous features. These findings underscore the importance of selecting appropriate $\alpha$ values to optimize the balance between homogeneity and heterogeneity within the BCL framework. Such adjustments should be tailored to the specific GAD method and the characteristics of the dataset in question, thereby enhancing the accuracy and robustness of anomaly detection.
\subsection{The Impact of $\lambda_{0}$ and $T$ (RQ6)}
\begin{figure}[htbp]
	\centering
	\includegraphics[width=1\linewidth]{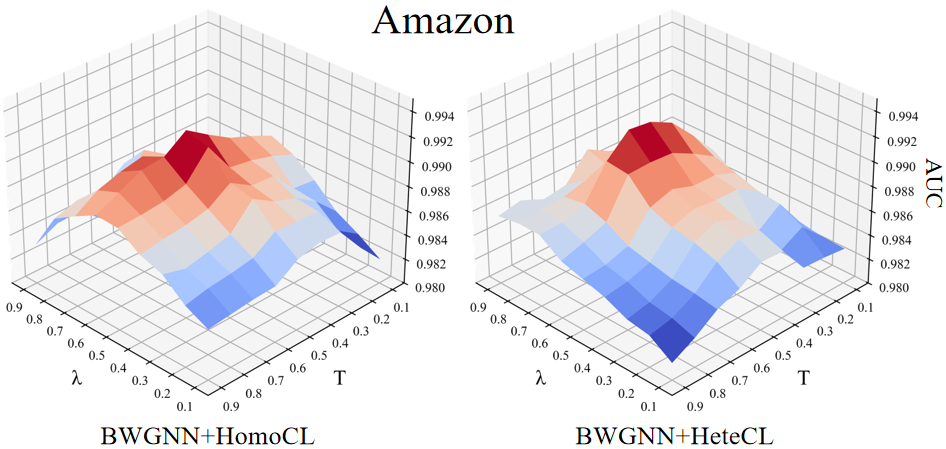} 
	\caption{The Impact of Training Scheduler}
	\label{fig:5}
\end{figure}
In this section, we investigate how the hyperparameters $\lambda_0$ and $T$ affect the performance of  BCL. Where $\lambda_0$ controls the initial number of training nodes, while $T$ controls the speed of introducing difficult nodes into the training process. We use BWGNN as the GAD and report its results on Amazon combined with HeteCL and HomoCL, respectively. From the results in Figure \ref{fig:4}, we can observe: (1) For Amazon, the performance tends to rise and then fall as $\lambda_0$ and $T$ increase. Too small or too small a pair of these parameters can degrade the performance of GAD. A very large $\lambda_0$ and $t$ will cause GAD to train mainly on easy subsets, resulting in the loss of information contained in difficult nodes. (2) Slightly different intervals of $\lambda_0$ can be found when focusing on homogeneity and heterogeneity. Specifically, when focusing on the direction of homogeneity, $\lambda_0$ between 0.5 and 0.6, $T$ at, 30\% to 60\%, the performance is relatively good. While in the direction of focusing on heterogeneity, it is between $\lambda_0$ between 0.4 and 0.5, $T$ between 40\% and 60\% that works relatively well. In addition, this characteristic varies depending on different datasets.

\section{Conclusion}

In this paper, we propose a Bi-directional Curriculum Learning approach, offering a novel and effective training paradigm for GAD. BCL employs a simple and effective training strategy to improve model performance through GAD's ability to handle both homogeneous and heterogeneous characteristics. We conducted extensive experiments on seven widely used datasets. The results demonstrate that BCL significantly outperforms other graph curriculum learning methods, leading to substantial improvements in the detection performance of existing GAD techniques.

%% The file named.bst is a bibliography style file for BibTeX 0.99c
\bibliographystyle{named}
\bibliography{ijcai25}

\end{document}